\documentclass[conference]{IEEEtran}
\IEEEoverridecommandlockouts
\usepackage{cite}
\usepackage{amsmath,amssymb,amsfonts}
\usepackage{algorithmic}
\usepackage{graphicx}
\usepackage{textcomp}
\usepackage{subcaption}
\usepackage{subfloat}
\usepackage{threeparttable}
\usepackage{xcolor}
\usepackage{multirow}
\usepackage{booktabs}
\usepackage{makecell}
\usepackage{amsthm}
\usepackage{algorithmic}
\usepackage{hhline}
\usepackage[T1]{fontenc}
\usepackage[ruled,linesnumbered]{algorithm2e}
\def\BibTeX{{\rm B\kern-.05em{\sc i\kern-.025em b}\kern-.08em
    T\kern-.1667em\lower.7ex\hbox{E}\kern-.125emX}}
\begin{document}
	%
\title{RL-LLM-DT: An Automatic Decision Tree Generation Method Based on RL Evaluation and LLM Enhancement
}
	%
	%
	%

\author{
    Junjie Lin,
    Jian Zhao,
    Lin Liu,
    Yue Deng,
    Youpeng Zhao,
    Lanxiao Huang,
    Xia Lin,
    Wengang Zhou,
    Houqiang Li
}
\maketitle
	
\begin{abstract}
Traditionally, AI development for two-player zero-sum games has relied on two primary techniques: decision trees and reinforcement learning (RL).
A common approach involves using a fixed decision tree as one player's strategy while training an RL agent as the opponent to identify vulnerabilities in the decision tree, thereby improving its strategic strength iteratively.
However, this process often requires significant human intervention to refine the decision tree after identifying its weaknesses, resulting in inefficiencies and hindering full automation of the strategy enhancement process.
Fortunately, the advent of Large Language Models (LLMs) offers a transformative opportunity to automate the process.
We propose RL-LLM-DT, an automatic decision tree generation method based on RL Evaluation and LLM Enhancement.
Given an initial decision tree, the method involves two important iterative steps.
\textbf{Response Policy Search}: RL is used to discover counter-strategies targeting the decision tree.
\textbf{Policy Improvement}: LLMs analyze failure scenarios and generate improved decision tree code.
In our method, RL focuses on finding the decision tree's flaws while LLM is prompted to generate an improved version of the decision tree.
The iterative refinement process terminates when RL can't find any flaw of the tree or LLM fails to improve the tree.
To evaluate the effectiveness of this integrated approach, we conducted experiments in a curling game.
After iterative refinements, our curling AI based on the decision tree ranks first on the Jidi\footnote{http://www.jidiai.cn/} platform among 34 curling AIs in total, which demonstrates that LLMs can significantly enhance the robustness and adaptability of decision trees, representing a substantial advancement in the field of Game AI.
Our code is available at https://github.com/Linjunjie99/RL-LLM-DT.
\end{abstract}
	
\section{Introduction}
The rise of artificial intelligence (AI) has brought about transformative changes in various domains, including sports and e-sports. 
In the realm of sports, AI has been employed to analyze player performance, optimize training regimens, and even simulate game strategies. 
E-sports, on the other hand, have seen AI as a natural extension of competitive gaming, where algorithms can compete against human players or assist in decision-making processes.
Reinforcement learning (RL), a subfield of machine learning, has emerged as a powerful tool for creating AI agents capable of learning optimal strategies through trial and error \cite{baier2018emulating,silva2018strategy}. 
In competitive environments, RL has demonstrated remarkable success, such as mastering complex board games like Go~ \cite{silver2016mastering,silver2017mastering} and achieving superhuman performance in video games such as Dota 2 \cite{berner2019dota} and StarCraft II \cite{vinyals2019grandmaster} and in card games like DouDiZhu \cite{zhao2023full} and GuanDan \cite{zhao2024danzero+}. 
The ability of RL to adapt and improve through experience makes it particularly well-suited for dynamic and unpredictable scenarios.

While RL-based AI has shown promise in competitive gaming, it often struggles with stability when facing unfamiliar opponents.
By contrast, the decision tree performs stably against diverse opponents and its structured essence brings better interpretability.
However, the decision tree requires extensive prior knowledge and otherwise is prone to errors in edge cases, leading to unexpected outcomes. 
Our former work \cite{lin2023mastering} combines the advantages of RL and decision tree.
We utilize RL to train a policy model against the developed decision tree to detect its flaws and manually refine the decision tree according to the detected flaws.
However, the reliance on human intervention in this process rendered the method non-automated, thereby limiting its scalability and practical applicability. 
Fortunately, the emergence of LLMs has demonstrated significant potential to automate and enhance this refinement process.

LLMs, such as OpenAI's GPT series \cite{wei2022emergent}, have demonstrated an unparalleled ability to understand and generate human-like text, making them ideal for tasks that require nuanced understanding and complex decision-making.
Recent studies, including but not limited to ReAct \cite{yao2022react}, SayCan \cite{ahn2022can}, Toolformer \cite{schick2024toolformer}, HuggingGPT \cite{shen2024hugginggpt} and WebGPT \cite{nakano2021webgpt}, have substantiated the viability of autonomous decision-making agents that are built upon a large language model foundation.
The emergence of code-oriented Large Language Models \cite{guo2024deepseek,hui2024qwen2} has significantly enhanced the robustness of code generation capabilities, thereby introducing a novel paradigm for their application. 
During the pre-training phase, LLMs assimilate extensive human experience and domain-specific knowledge, particularly within the coding domain, thus establishing a robust foundation for generating code capable of addressing intricate tasks. 
These agents can take task descriptions as inputs, subsequently generating corresponding decision tree code.

In this paper, we integrate RL and LLMs to develop high-performance and robust AI decision trees for curling game. 
This hybrid approach employs RL to identify the decision tree's shortcomings and utilizes LLMs to refine and improve the decision tree. 
We establish the decision tree as the built-in opponent within the curling game environment and train a neural model to surpass it through RL. 
Subsequently, we leverage LLMs to analyze the game trajectories where the decision tree loses to neural models, gaining deeper insights into its vulnerabilities. 
This analysis enables us to refine the decision tree to enhance its performance. 
Additionally, LLMs are employed to generate the strategy code directly from the improved decision tree, ensuring consistency and accuracy in implementation. 

Through this innovative integration of LLMs, decision trees, and reinforcement learning, we aim to push the boundaries of AI in sports and e-sports, offering a robust and versatile framework for future research and application.

\section{Related Work}
\subsection{Sports Game AI}
The integration of artificial intelligence (AI) into sports strategy generation has garnered substantial academic and practical interest in recent years. 
A significant advancement in this domain is the Google Research Football Environment \cite{kurach2020google}, a physics-based 3D simulation designed for reinforcement learning in football. 
This environment facilitates the training of AI agents by replicating real-world football scenarios, featuring eleven distinct game scenes and an internal opponent AI with adjustable difficulty levels. 
Notably, this initiative has attracted the attention of Manchester City, one of the world's leading football clubs. 
Manchester City has collaborated with Google to explore the application of football AI in enhancing the training of professional football players, underscoring the potential of AI in refining tactical decision-making and performance.
In basketball, NetEase has developed basketball AI for its mobile game Street Basketball~ \cite{jia2020fever}, which has demonstrated proficiency in fundamental skills such as passing, shooting, and penetration, as well as advanced tactical maneuvers such as pick-and-roll plays, team defense, and strategic passing following penetrations. 

The domain of intellectual sports, such as chess and card games, has also witnessed remarkable achievements in AI development. 
For instance, Google's AlphaGo \cite{silver2016mastering} and its successor AlphaGo Zero \cite{silver2017mastering} have achieved landmark victories over professional Go players, revolutionizing the field of AI-assisted strategy development. 
Many professional Go players now incorporate Go AI into their training regimens to refine their skills and analyze game strategies. 
Similarly, Suphx  \cite{li2020suphx}, a mahjong AI that integrates supervised learning, reinforcement learning, and specialized techniques, has surpassed 99.99\% of players on the Japanese mahjong platform Tenhou and has outperformed top-ranked masters. 
Additionally, DouZero \cite{zha2021douzero}, an AI system for the card game DouDiZhu, has demonstrated performance levels comparable to top human players, highlighting the versatility of AI in competitive gaming.

In the realm of e-sports, AI developed through deep reinforcement learning have exhibited superior performance compared to elite human professional players. 
A notable example is AlphaStar \cite{vinyals2019grandmaster}, an AI designed to master the complex Real-Time Strategy (RTS) game StarCraft II. 
AlphaStar achieved a milestone by defeating a top human player in an exhibition match, adhering to limited Actions Per Minute (APM) conditions to ensure fairness. 
Another significant achievement is OpenAI Five \cite{berner2019dota}, an AI system for the Multiplayer Online Battle Arena (MOBA) game Dota 2, which triumphed over a championship-level team. 
These successes have not only showcased the capabilities of AI in competitive gaming but have also inspired human players to engage with and challenge game AI, fostering a symbiotic relationship between human and machine in the pursuit of strategic excellence.

\subsection{LLM Agents}

Large Language Model (LLM) agents have demonstrated remarkable capabilities in complex game environments, showcasing their potential in understanding, reasoning and decision making. 
ChessGPT \cite{feng2024chessgpt}, for instance, integrates policy learning with language modeling by leveraging data from both game and language datasets to enhance performance in Chess games. 
Voyager \cite{wang2023voyager}, on the other hand, highlights the ability of LLM agents to navigate and perform tasks within the MineDojo \cite{fan2022minedojo} environment, a complex open-world simulation. 
In the realm of Real-Time Strategy (RTS) games, TextStarCraft \cite{ma2023large} introduces a natural language interface that enables LLMs to engage in StarCraft II. 
By employing the Chain-of-Summarization method, TextStarCraft facilitates efficient reasoning and decision making, allowing the agent to process complex game states and generate coherent strategies. 
Pokéllmon \cite{hu2024pok} represents another significant advancement, introducing an environment that allows LLMs to participate in Pokémon battles. 
The agent developed by Pokéllmon achieves human-level performance by consuming instant feedback to iteratively refine its policy. 
Additionally, it incorporates mechanisms for retrieving external knowledge to mitigate hallucination, a common challenge in LLM-based systems. 
This dual approach enhances the agent's robustness and adaptability, enabling it to perform competently in dynamic and knowledge-intensive game scenarios.
Furthermore, Cradle \cite{tan2024towards} introduces a multi-modal agent capable of perceiving the game screen, analyzing game instructions, generating action plans, and directly controlling characters through mouse/keyboard operations in the renowned 3D action-adventure game Red Dead Redemption 2 (RDR2) based on the multi-modal LLM GPT-4V. 

With the increasing robustness of LLMs in generating code, direct generation of executable codes or scripts becomes feasible. 
Researchers from UIUC and Apple propose the general framework CodeAct \cite{wang2024executable}, which allows LLMs to generate executable Python code as their action. 
Compared to JSON and pre-formatted text formats, code inherently supports control and data flow, enabling intermediate results to be stored as variables for reuse. 
A single piece of code can combine multiple tools to perform complex logical operations (e.g., if statements and for loops), unlocking the potential of pre-trained programming knowledge of large models to handle complex tasks.
Eureka \cite{ma2023eureka} is a general-purpose reward design algorithm driven by coding large language models and contextual evolutionary search. 
Without prompt engineering or human intervention for any specific task, Eureka enables human-level reward generation across a wide range of robots and tasks.

\section{Background}
In this section, we briefly introduce the curling game environment and review the concept and formulation of reinforcement learning.
\subsection{Curling Game Environment}
\begin{figure}[t]
	\centering
	\includegraphics[width=1.0\columnwidth]{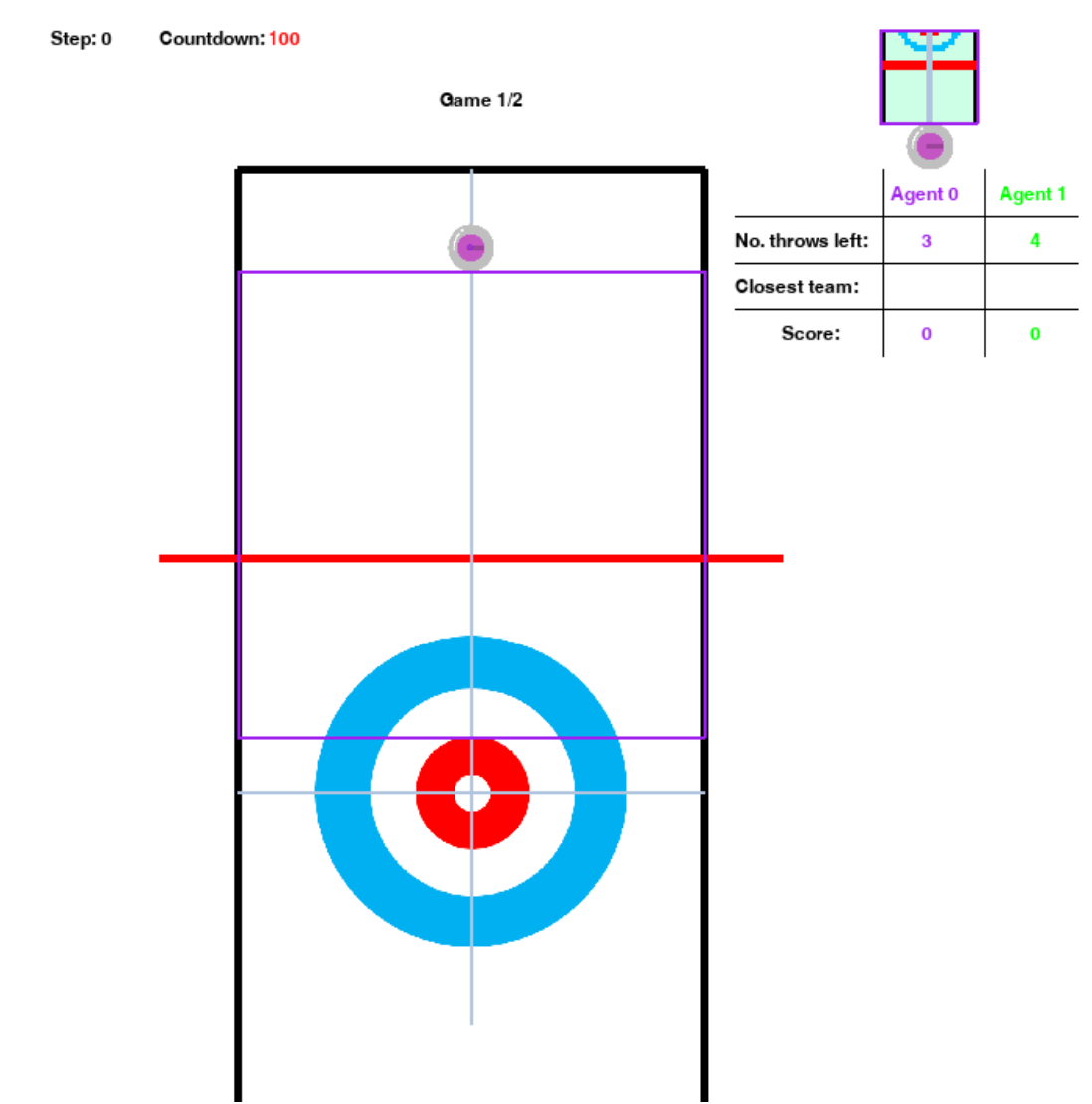}
	\caption{An illustration of the virtual environment of curling game.}
	\label{fig:init}
\end{figure}

We choose the curling game environment developed by Jidi platform for our study.
The visualization of the curling game environment can be seen in Figure~\ref{fig:init}.
\subsubsection{Game Rules}

In the curling game environment, two teams compete against each other, taking turns to throw stones.
The detailed rules of the curling game environment are as below:
\begin{itemize}
    \item Each team controls four elastic spherical stones with the same mass and radius.
    \item The agents of both teams take turns to throw the stones to the target point in the center of the field, and each agent has four chances to throw in each game.
    \item Stones can collide with each other and against walls, but they will lose a certain amount of velocity after collision.
    \item The agent can no longer exert force on the stones after it has passed the red line.
    \item  The agent's field of view is limited to a matrix area of 30*30 facing forward.
    \item Two games in total and swap the order of serve in the second game.
    \item At the end of each game, the score will be calculated, and one point will be awarded for each curling stone of one team that is closer to the center area than all the curling stones of the other team, and only one team will score points in each game.
    \item At the end of the two games, the winner will be determined based on the total score of the two games, with the team with the highest score being the winner.
    \item  When the two games are over, the environment ends.
\end{itemize}
\subsubsection{Game Pre-process}\label{preprocess}

Observation pre-process: as mentioned in the game rules section, agents can only receive a 2-dimension image-like observation.
However, truly useful information is about the positions of the current stones on the ice.
Therefore, we utilize template matching to calculate the coordinates of the stones.
Both decision tree AI and neural model AI take the concrete coordinates information as input rather than abstract image-like observation.

Action pre-process: before a stone crosses the red line, agents can exert a force with a specific angle and magnitude.
In other words, the trajectory of a stone after it crosses the red line is determined by its position and velocity.
Therefore, there is no need to change the angle and magnitude of the exerted force at each step.
Instead, selecting an initial position and throwing the stone with a fixed angle and fixed magnitude force can reduce the complexity while maintaining flexibility.

\subsubsection{Game Platform}
Jidi is an online competition platform that provides users with massive environments, high-quality competitions, real-time discussions and fair algorithm rankings.
The platform boasts high levels of openness and source availability in areas such as agent algorithms, arena competitions, and forum communities, offering a channel for exchange and sharing among researchers, students, and industry professionals in the field of game decision-making both domestically and internationally.
We submit our curling AI to this platform for testing our AI's ability.

\subsection{Reinforcement Learning}
Reinforcement learning \cite{sutton2018reinforcement} (RL) is a type of machine learning that involves an agent learning to make decisions by interacting with an environment. 
The goal of the agent is to learn a strategy for selecting actions that maximizes some notion of cumulative reward over time.
\subsubsection{Markov Decisioin Process}
A Markov Decision Process (MDP) \cite{sutton2018reinforcement} is a mathematical framework used to model decision-making in situations where outcomes are influenced both by random factors and by the actions of a decision maker.
MDPs are particularly useful in reinforcement learning, where they provide a way to formalize the problem of learning from interactions with an environment to achieve a goal.
A MDP can be defined as a tuple $(\mathcal{S},\mathcal{A},\mathcal{P},\mathcal{R},\gamma)$, in which $\mathcal{S}$ denotes the set of possible states, $\mathcal{A}$ denotes the set of possible actions, $\mathcal{P}$ denotes the transition probability function, $\mathcal{R}$ denotes the reward function, and $\gamma$ denotes the discount factor that determines the importance of future rewards.
At each step $t$, the agent takes an action $a_t\in \mathcal{A}$ according to the current state $s_t\in \mathcal{S}$.
Then the environment state transforms to $s_{t+1}$ according to the probability of $\mathcal{P}(s_{t+1}|s_t,a_t)$.
Meanwhile, the agent receives a reward signal $r_t=\mathcal{R}(s_t,a_t,s_{t+1})$.
The accumulated reward is defined as: 
\begin{equation}\label{eq1}
R_t = \sum_{t^{\prime}=t}^{T} \gamma^{t^{\prime}-t}r_{t^{\prime}}.
\end{equation}\label{eq2}
The expected cumulative reward of a certain policy $\pi$ for a given
state $s$ is calculated as follows:
\begin{equation}\label{eq3}
	\mathcal{R}^{\pi}(s)=\mathbb{E}_{a_t\sim \pi,s_{t+1}\sim \mathcal{P}}[\sum_{t=0}^{\infty} r_{t}|s_{0}=s].
\end{equation}
The primary objective of RL is to find an optimal policy that maximizes the expected cumulative reward:
\begin{equation}\label{eq3}
J(\pi) = \mathbb{E}_{s_0 \sim \mathcal{P}}[\mathcal{R}^{\pi}(s_0)].
\end{equation}

\subsubsection{Proximal Policy Optimization}
PPO \cite{schulman2017proximal} is a popular algorithm in reinforcement learning, particularly in the field of deep reinforcement learning. 
It is introduced as an improvement over previous policy gradient methods \cite{schulman2015trust}, aiming to provide a balance between ease of implementation, sample efficiency, and computational complexity.
PPO is a policy-based algorithm and adopts an actor-critic framework.
The value loss of PPO is defined as:
\begin{equation}\label{eq4}
    \mathcal{L}_{value}=\mathbb{E}[(r_t+\gamma V_{\theta}(s_{t+1})-V_{\theta}(s_t))^2].
\end{equation}
In order to alleviate the effect of off-policy learning, PPO utilizes a clipped policy loss:
\begin{equation}\label{eq5}
\begin{split}
    \mathcal{L}^{clipped}_{policy}=&-\mathbb{E}[\min(ratioA^{\pi_{old}}(a|s),\\
    &\text{clip}(1-\epsilon,ratio,1+\epsilon)A^{\pi_{old}}(a|s))],
\end{split}
\end{equation}
\begin{equation}\label{eq6}
\text{clip}(1-\epsilon, ratio, 1+\epsilon) = \left\{
\begin{array}{ll}
1+\epsilon,   & ratio > 1+\epsilon\\
ratio,      & 1-\epsilon\leq ratio\leq 1+\epsilon\\
1-\epsilon,   & ratio < 1-\epsilon \\
\end{array}
\right. ,
\end{equation}
where $ratio=\frac{\pi_\theta(a|s)}{\pi_{old}(a|s)}$, $A^{\pi_{old}}(a|s)$ means the advantage estimation which is calculated by Generalized Advantage Estimation (GAE)  \cite{schulman2015high}.
PPO utilizes entropy loss to encourage exploration:
\begin{equation}\label{eq7}
    \mathcal{L}_{entropy}=\mathbb{E}_s[\sum_{a\in\mathcal{A}}\pi_{\theta}(a|s)\log\pi_{\theta}(a|s)].
\end{equation}
Overall, the total loss of PPO is calculated as:
\begin{equation}\label{eq8}
    \mathcal{L}_{PPO}=\mathcal{L}_{policy}^{clipped}+c_v\mathcal{L}_{value}+c_e\mathcal{L}_{entropy},
\end{equation}
in which $c_v$ is the value coefficient and $c_e$ is the entropy coefficient.

\section{Method}
In this section, we introduce our hybrid method to build decision tree with robust and outstanding performance.
The reasoning and coding capability of LLM helps generate decision tree policy and decision tree code, while RL contributes to finding the flaws of the decision tree.
Following is a detailed discussion for our framework.

\subsection{Framework}
\begin{figure*}[t]
	\centering
	\includegraphics[width=2.0\columnwidth]{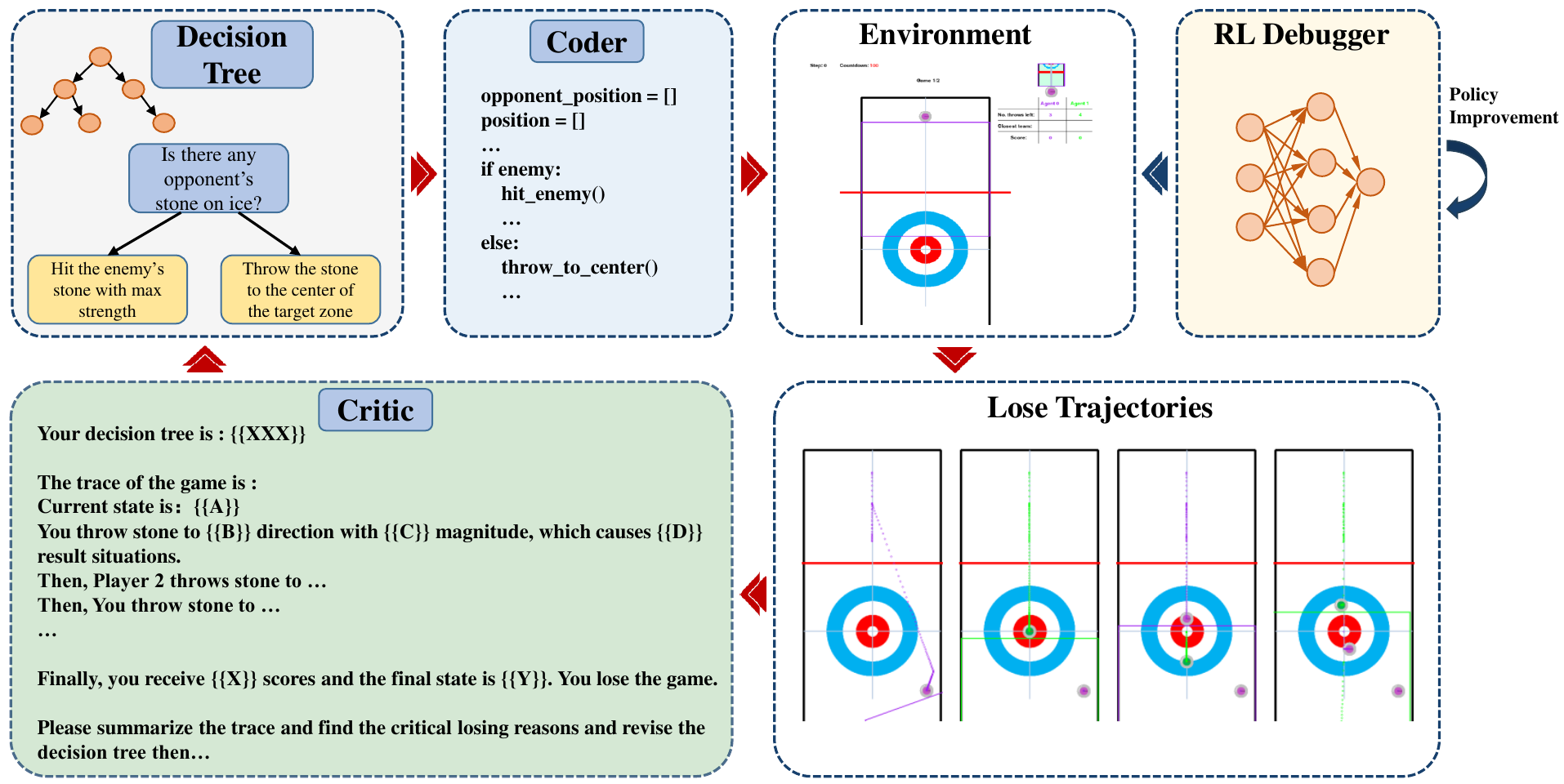}
	\caption{An illustration of the overall framework.}
	\label{fig:framework}
\end{figure*}
The proposed framework, illustrated in Figure~\ref{fig:framework}, is composed of three pivotal modules: the Large Language Model (LLM) Coder, the Reinforcement Learning (RL) Debugger, and the LLM Critic. 
This framework is designed to iteratively refine decision tree tactics within the context of the curling game environment.

The LLM Coder module is responsible for generating Python scripts that implement the specified decision tree tactic. 
Given the LLM's limited prior knowledge of the curling game environment, the prompt provides detailed descriptions and uses of the available interfaces. 
This contextual information is crucial for the LLM to generate executable decision tree code that accurately reflects the desired tactic.

Once the decision tree code is generated, it is passed to the RL Debugger module. 
Here, a customized deep neural model is trained using a deep reinforcement learning algorithm to compete against the decision tree code. 
During the training process, the deep neural network iteratively improves its performance through interactions with the curling game environment. 
If the decision tree code is defeated by the deep neural model, detailed records of the lost games are collected. 
These records include essential information, such as state transitions and actions taken during gameplay.

Subsequently, the information about lost games is passed to the LLM Critic module. 
The LLM Critic analyzes the game information to summarize the reasons behind the decision tree's defeat. 
This analysis involves identifying patterns, errors, or inefficiencies in the decision tree tactic that led to its failure.
Based on this analysis, the LLM Critic designs an improved version of the decision tree tactic. 
The improvement process leverages both the previous decision tree information and the feedback from the failure of previous games, which aims to address the identified flaws and enhance the decision tree's performance.

The improved decision tree tactic is then re-evaluated within the framework. 
The refined decision tree code is passed back to the RL Debugger module, where the deep neural model is retrained to compete against it. 
This loop continues iteratively, allowing the decision tree tactic to be refined progressively.
The iterative refinement process terminates when one of the following conditions is met:
The RL Debugger module is unable to identify any further flaws in the decision tree tactic.
The LLM Critic module is unable to propose any further improvements to the decision tree tactic.

\subsection{LLM Coder}
The coder module is utilized to generate the executable Python code to implement a given decision tree tactic.
As the rules and APIs of the curling game environment are specific, they need to be clearly defined in the prompt.
We use $I_{rules}$ to denote information about the curling game environment and use $I_{APIs}$ to represent available environment interfaces.
Then executable decision tree code can be obtained from the LLM Coder:
\begin{equation}\label{eq7}
    C=LLM_{coder}(DT,I_{rules},I_{APIs}),
\end{equation}
where the decision tree code $C$ is generated based on given decision tree tactic $DT$, rules information $I_{rules}$ and interfaces information $I_{APIs}$.
The generated code $C$ is evaluated in the RL Debugger module.
A customized deep neural model will be trained against it.
Also, in order to check the generated code's robustness and ability in the curling game, we submit the code to Jidi platform.

\subsection{RL Debugger}
\begin{algorithm}[h]
    \caption{Actor Process}
    \label{alg:algorithm1}
    \begin{algorithmic}[1]
    \STATE Initialize local policy model $M$ and local buffers $D$;
    \FOR{iteration = 1, 2, 3, ...}
        \STATE Synchronize policy model $M$ with the learner \\ process and reset the game environment;
        \FOR{t = 0, 1, 2, ..., T}
            \STATE Calculate $a_t,v_t,\pi(a_t|s_t)$ according to $s_t$ \\ through
            model $M$;
            
            \STATE Perform $a_t$ and observe $s_{t+1}$ and reward $r_t$;
            
            \STATE Store $(s_t,a_t,r_t,v_t,\pi(a_t|s_t))$ to buffer $D$;
        \ENDFOR
        \STATE $v_{T+1}\leftarrow0$;
        \STATE $d_{T+1}\leftarrow0$;
        \FOR{$t$ = $T$, $T$-1, ..., 0}
            \STATE $d_t\leftarrow r_t+\gamma v_{t+1}-v_t+\gamma\lambda d_{t+1}$;
            \STATE $r_t\leftarrow d_t+v_t$ and update $r_t$ in $D$;
        \ENDFOR
        \STATE Send data $(s_t,a_t,r_t,v_t,\pi(a_t|s_t))$ from buffer $D$ to shared buffer $B$;
    \ENDFOR
    \end{algorithmic}
\end{algorithm}
\begin{algorithm}[t]
    \caption{Learner Process}
    \label{alg:algorithm2}
    \begin{algorithmic}[1]
    \STATE Initialize local policy model $M$;
    \FOR{iteration = 1, 2, 3, ... until convergence}
    \IF{Shared buffer $B$ is half full}
        \STATE Sample a batch of $(s_t,a_t,r_t,v_t,\pi(a_t|s_t))$ from \\ $B$ and the batch size is accurately the half size \\ of $B$;
        \STATE Update model $M$ with value loss and clipped \\
        policy loss in PPO and learning rate $\psi$;
    \ENDIF
    \ENDFOR
    \end{algorithmic}
\end{algorithm}
In the RL Debugger module, a deep neural model is trained against the generated decision tree code $C$ in the curling game environment.
Deep RL algorithms can be freely applied in this module. 
In order to accelerate the training process, we utilize the distributed version of PPO as our training algorithm.
We adopt multiple actor processes to interact with the curling game environment for collecting samples in the replay buffer, which is detailed in the Algorithm~\ref{alg:algorithm1}.
As for the single learner process summarized in Algorithm~\ref{alg:algorithm2}, the deep neural model is optimized using the PPO loss on the trajectories data from actor processes.
In order to prevent the sampling policy from deviating far from the learning policy, the updated deep neural model is synchronized periodically with the model in the actor process. 
When the win rate of the neural model stabilizes, a customized neural model against the generated decision tree code is obtained:
\begin{equation}\label{eq8}
    M_{C}=RL_{debugger}(C,Env_{curling}).
\end{equation}
The game information about the customized neural model $M_C$ against the generated decision tree code $C$ from the LLM Coder is recorded as:
\begin{equation}\label{eq9}
    I_{trace}=Env_{curling}(M_C,C),
\end{equation}
which then is passed to the LLM Critic module as a feedback signal to help analyze the flaws of the current decision tree.

\subsection{LLM Critic}
The LLM Critic module aims to generate an improved version of the decision tree tactic.
It takes the decision tree description, the corresponding implemented code and the game feedback information as input.
The LLM Critic evaluates the drawbacks of the decision tree and analyzes the cause of why the neural model beats the decision tree code.
The improved version of the decision tree is directly derived from the critic module:
\begin{equation}\label{eq10}
    DT_{new}=LLM_{critic}(DT_{old},I_{rules},I_{APIs},C,I_{trace}).
\end{equation}
Similar to in the LLM Coder module, curling game rules $I_{rules}$ and available interfaces $I_{APIs}$ are detailed in the prompt for the LLM Critic.
The new version of the decision tree will run in our framework for a new loop to be iteratively refined, increasing the robustness and ability of the decision tree.

\section{Experiments}
In this section, we initiate a simple decision tree tactic as a start and progressively refine it in our framework until the LLM can't give a better version of the decision tree.
Within the process, the decision tree becomes stronger and stronger and it takes a longer time for the RL debugger to train a deep neural model to beat it.
We submit our different versions of the decision tree code to the Jidi platform to compete against other users' curling AI.
The final version of our curling AI ranks first on the agents' PK platform, showing the effectiveness of our method and the ability of LLMs to design curling game strategies and implement corresponding Python scripts.

\subsection{The Efficacy of the Framework}
\begin{table}[t]
    \centering
    \setlength{\tabcolsep}{0.04\linewidth}{
    \renewcommand{\arraystretch}{2.0}{
    \begin{tabular}{c|cccc}
    \hline
    version&  Tree \uppercase\expandafter{\romannumeral1}&  Tree \uppercase\expandafter{\romannumeral2}& Tree \uppercase\expandafter{\romannumeral3}& Human Design\\
    \hline
    rank&9&2&1&3\\
    score&0.47&0.90&0.93&0.80\\
    \hline
    \end{tabular}
    }
    }
    \caption{The performance of three LLM generated and refined decision trees and a human designed decision tree on Jidi platform.}
    \label{tab:rank}
\end{table}

\begin{figure}[t]
    \centering
    \begin{subfigure}[b]{0.45\textwidth}
        \includegraphics[width=\textwidth]{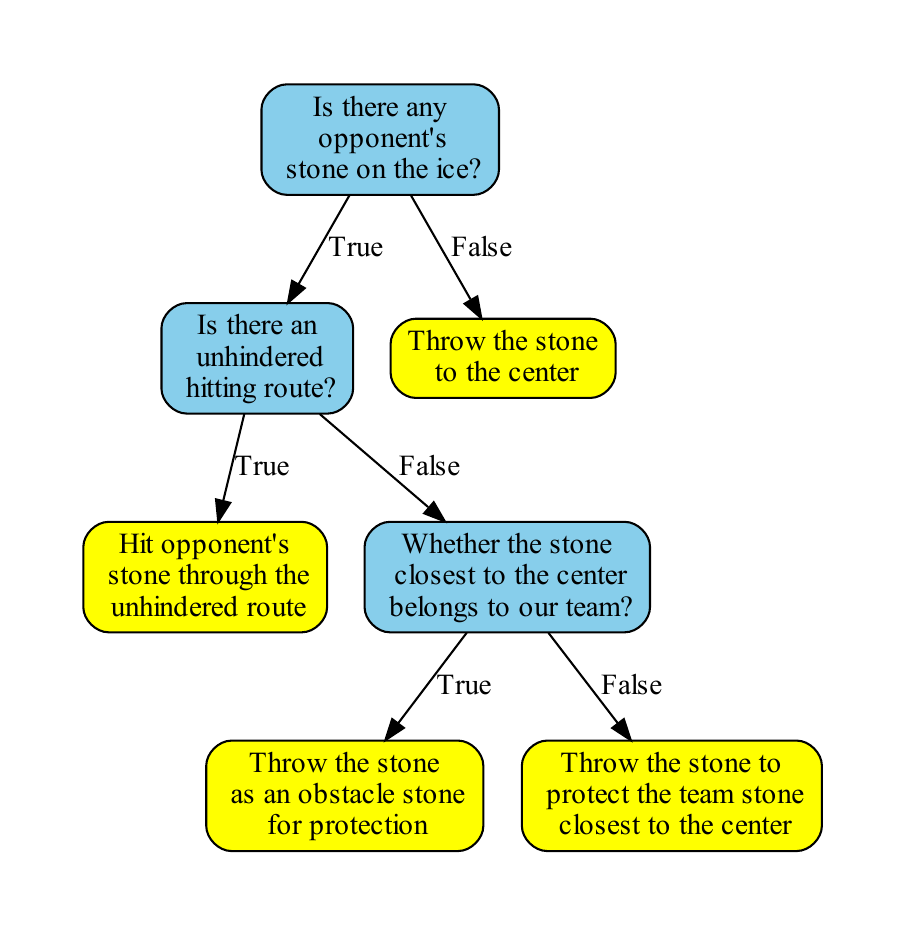}
        \caption{Final Decision Tree Generated by human.}
        \label{fig:tree3}
    \end{subfigure}
    \hfill
    \begin{subfigure}[b]{0.45\textwidth}
        \includegraphics[width=\textwidth]{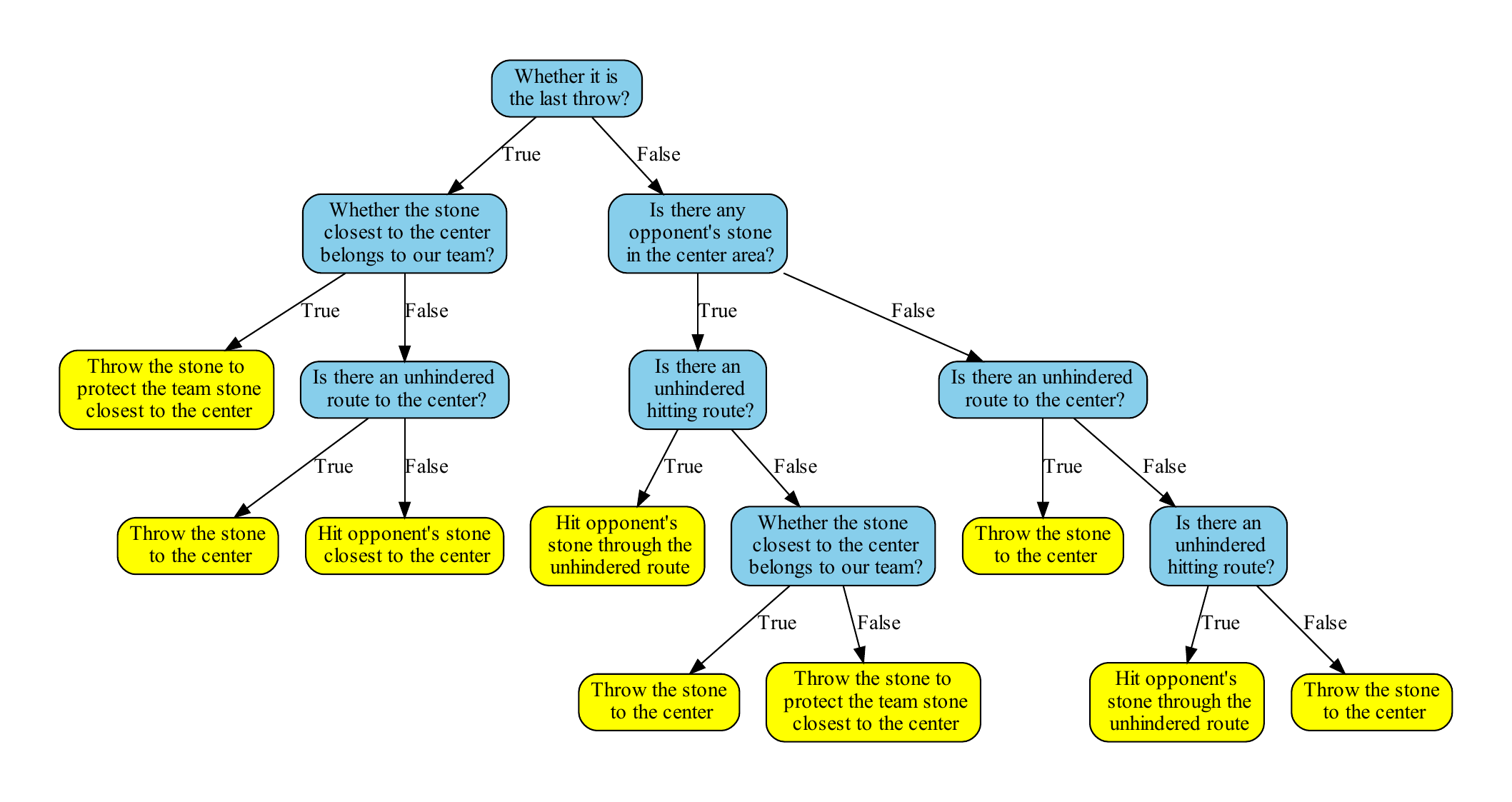}
        \caption{Final Decision Tree Generated by LLM.}
        \label{fig:tree4}
    \end{subfigure}
    \caption{The final decision tree developed by human in our former work and by LLM in this work.}
    \label{fig:tree_contrast}
\end{figure}

As evidenced by Table~\ref{tab:rank}, our iteratively refined decision tree curling AI demonstrates progressively superior performance on Jidi platform. 
The third iteration of the refined decision tree achieves the top rank on the platform, surpassing the curling AI of any other team, thereby validating the efficacy of our methodology. 
A total of 34 teams submitted their customized curling AI for evaluation.
Despite the diverse and unpredictable strategies employed by the opponents, our best decision tree curling AI attained an exceptionally high evaluation score of 0.93 out of 1 when competing against unseen opponent curling AIs, higher than the best decision tree version in our former work. 
This outcome reveals the robust performance of the generated decision tree curling AI.
As illustrated in Figure~\ref{fig:tree_contrast}, compared to the final decision tree refined by humans, the final decision tree generated by LLM shares a more complicated structure, which takes the remaining stones' number into account and prioritizes attack or protection according to the game situation.
It tends to hit the opponent's stone in the early stage of the curling game while prioritizing protection when it comes to the last throw.
Also, the decision tree generated by LLM tends to ignore opponent's stone away from the center area while the tree does not.
By incorporating LLM, we successfully develop a robust decision tree with high performance for the curling game, even better than the version generated by human in our former work.

\subsection{The Efficacy of LLM Coder}
For a given decision tree tactic, the LLM Coder module generates the corresponding Python code. 
At the outset of the iterative process, we initialize a rudimentary decision tree tactic and request the LLM Coder to translate this tactic into Python scripts. 
\begin{figure}[t]
	\centering
	\includegraphics[width=1.0\columnwidth]{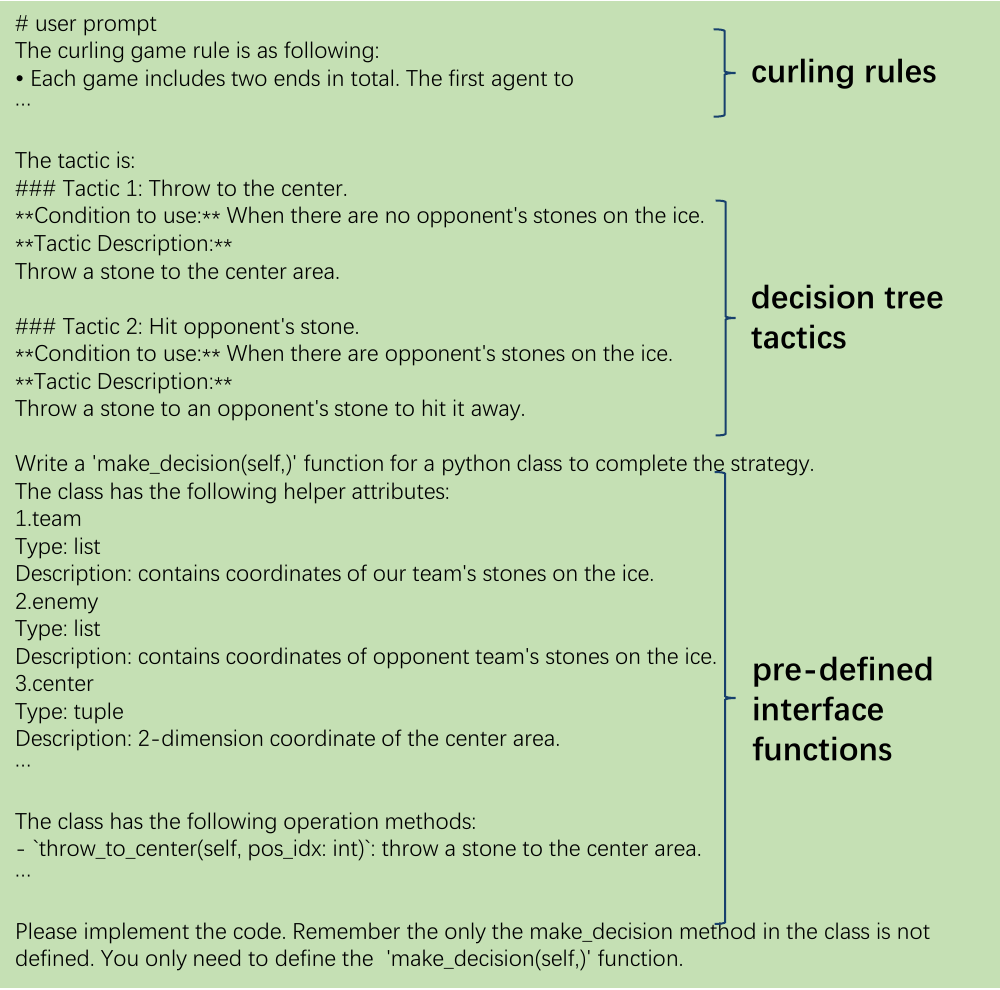}
	\caption{An LLM Coder prompt example.}
	\label{fig:coder1prompt}
\end{figure}
\begin{figure}[t]
	\centering
	\includegraphics[width=1.0\columnwidth]{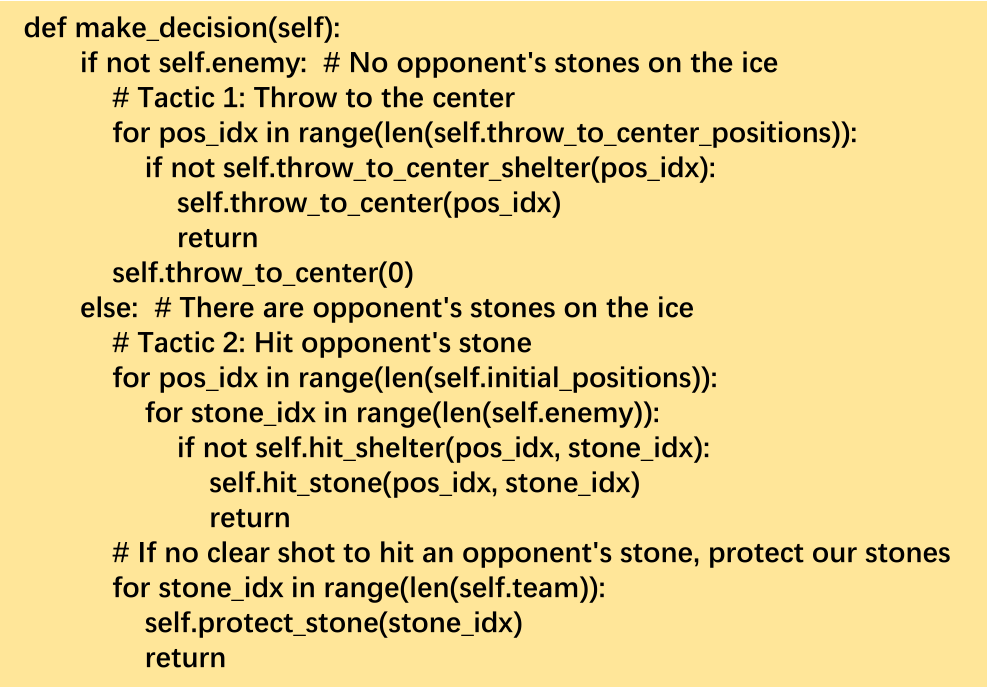}
	\caption{LLM Coder output for the initial decision tree.}
	\label{fig:coder1answer}
\end{figure}
An illustrative user prompt example is presented in Figure~\ref{fig:coder1prompt}, with the corresponding initial decision tree output showcased in Figure~\ref{fig:coder1answer}. 
Remarkably, the LLM Coder not only faithfully implements the Python code in accordance with the provided decision tree tactic description but also subtly enhances the decision tree by incorporating a protective operation, which was not explicitly demanded by the tactic. 
This enhancement underscores the LLM Coder's capability to autonomously refine and fortify the generated code.

\begin{figure}[t]
	\centering
	\includegraphics[width=1.0\columnwidth]{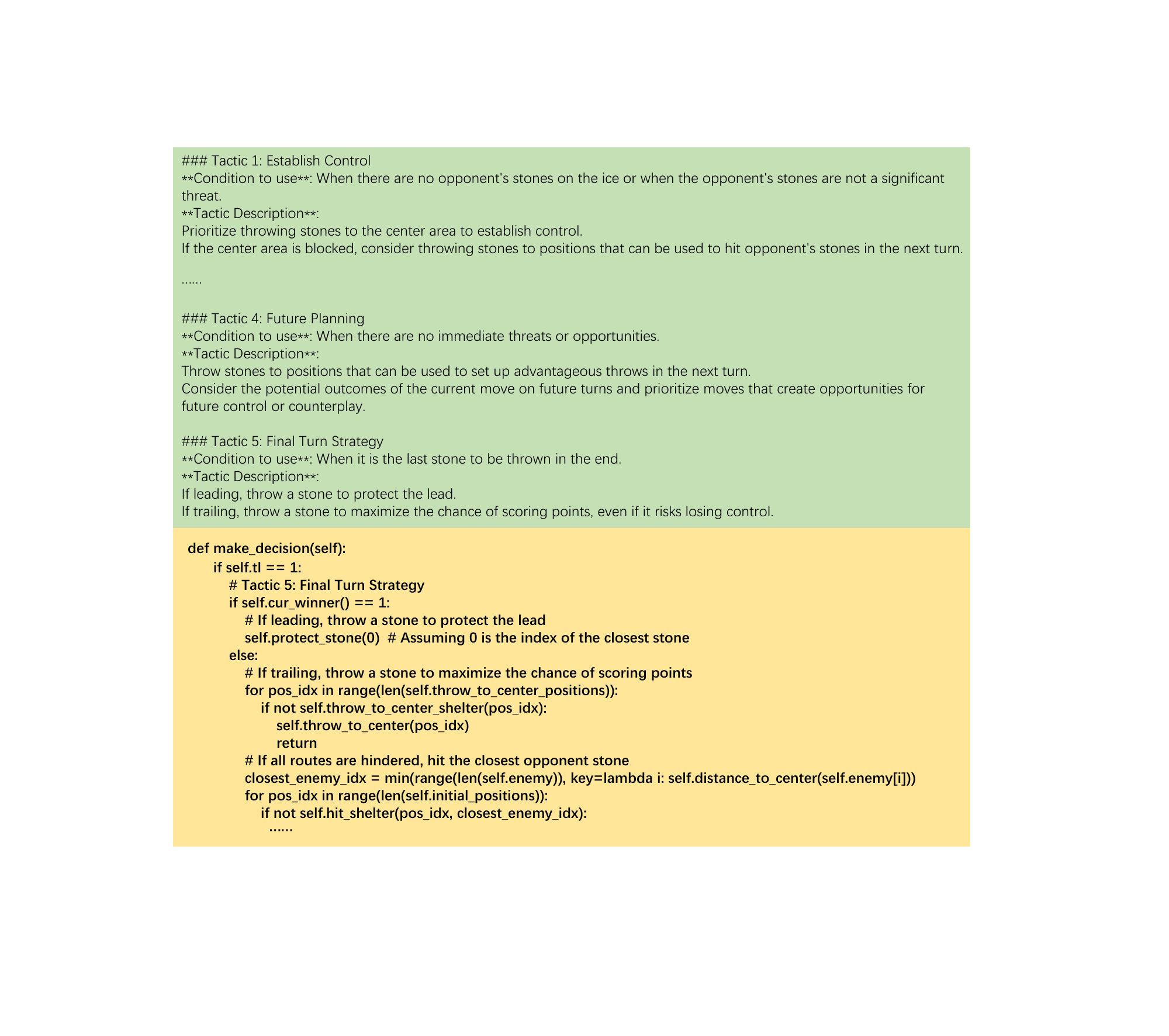}
	\caption{Even the refined decision tree tactic is not very clear, the LLM Coder still generates corresponding executable Python scripts.}
	\label{fig:robust}
\end{figure}
The LLM Coder exhibits remarkable consistency and robustness, as evidenced by Figure~\ref{fig:robust}. 
As the decision tree tactic is progressively refined through our method, it becomes increasingly intricate and challenging to implement, particularly when the tactic descriptions generated by the LLM Critic grow more ambiguous. 
Despite these escalating complexities, the LLM Coder maintains its functionality, consistently producing executable Python scripts. 
This resilience highlights the LLM Coder's robust coding proficiency and its adaptability to evolving and complex tactical requirements.

\subsection{The Efficacy of RL Debugger}
We apply distributed PPO to train a customized deep neural model against the decision tree code in the curling game environment.
As discussed in section~\ref{preprocess}, the policy model necessitates the incorporation of stone coordinates within its input domain. 
Furthermore, to augment the granularity of game-related information, the number of remaining stones for both our team and the opponent is integrated into the policy model's input. 
Due to the limited control accuracy of the applied force, the continuous action space of the curling game is systematically discretized through uniform sampling. 
This discretization enables the policy model to generate a discrete probability distribution as its output.
The architecture of our policy model is predicated upon a Multi-Layer Perceptron (MLP) framework. 
The input of the model comprises the coordinates of specific stones and the tally of remaining stones, rather than abstract graphical observations. 
This design ensures that the policy model is directly informed by the essential spatial and quantitative attributes of the game state, thereby enhancing its decision-making capabilities.

\begin{figure}[t]
    \centering
    \includegraphics[width=1.0\columnwidth]{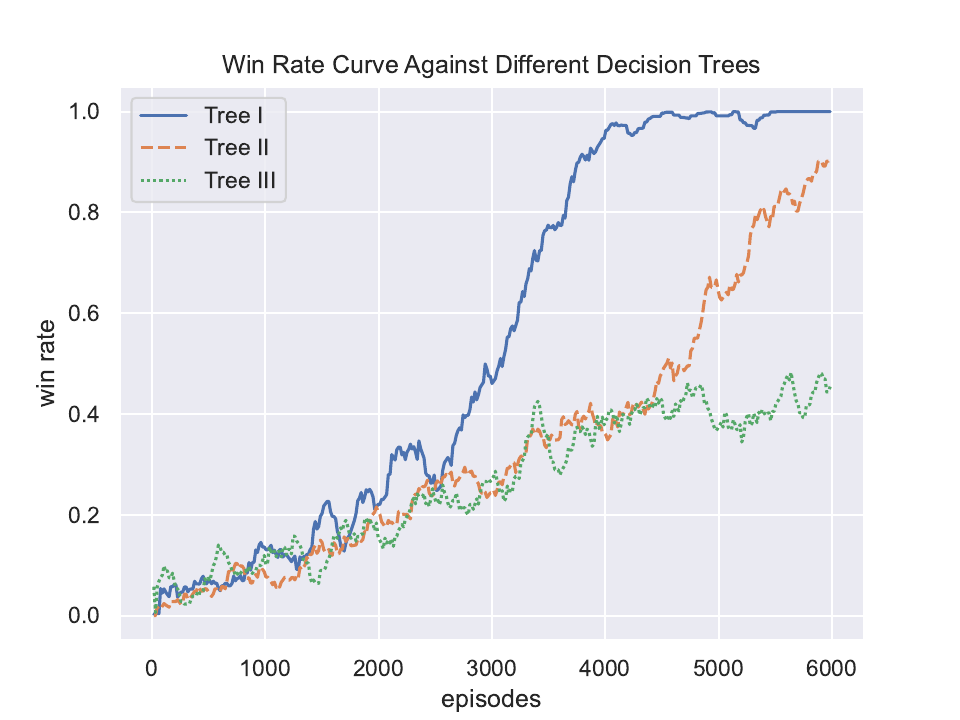}
    \caption{The winning rate of the policy model during training against three versions of the LLM generated and refined decision trees.}
    \label{fig:exp}
\end{figure}

The winning rate graph for the policy model, trained against various decision tree versions, is depicted in Figure~\ref{fig:exp}.
As the decision tree undergoes continuous refinement, it becomes progressively more challenging and time-consuming for the RL Debugger to train a customized deep neural model to identify its shortcomings and beat it. 
This trend underscores the escalating strength and robustness of the iteratively improved decision tree. 

\subsection{The Efficacy of LLM Critic}
\begin{figure}[t]
	\centering
	\includegraphics[width=1.0\columnwidth]{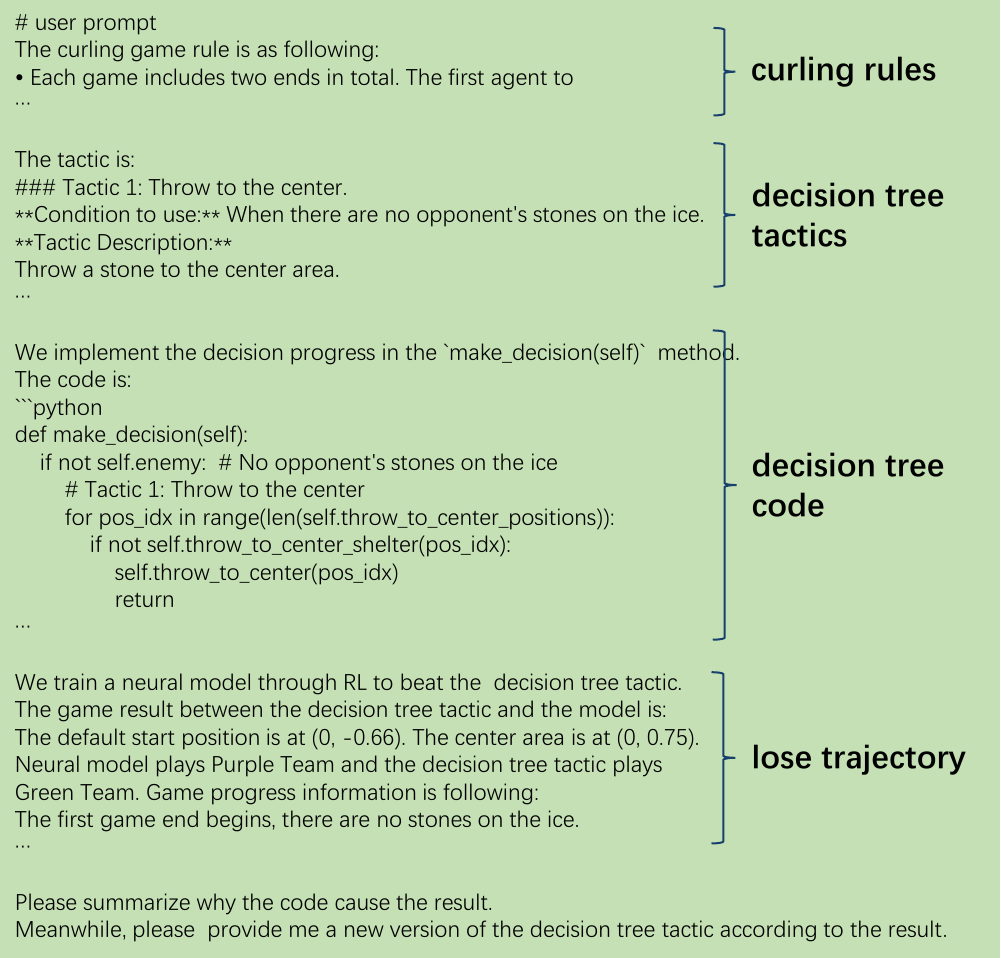}
	\caption{An illustration about the LLM Critic prompt.}
	\label{fig:criticprompt}
\end{figure}
As illustrated in Figure~\ref{fig:criticprompt}, the LLM Critic ingests the rules of the curling game, the description of the decision tree, the corresponding executable Python code implemented by the LLM Coder, and the trajectory of the lost game as input. 
The LLM Critic is tasked with diagnosing the reasons behind the decision tree code's failure in the curling game against the customized neural model and is required to generate an improved version of the decision tree tactic.

\begin{figure}[t]
	\centering
	\includegraphics[width=1.0\columnwidth]{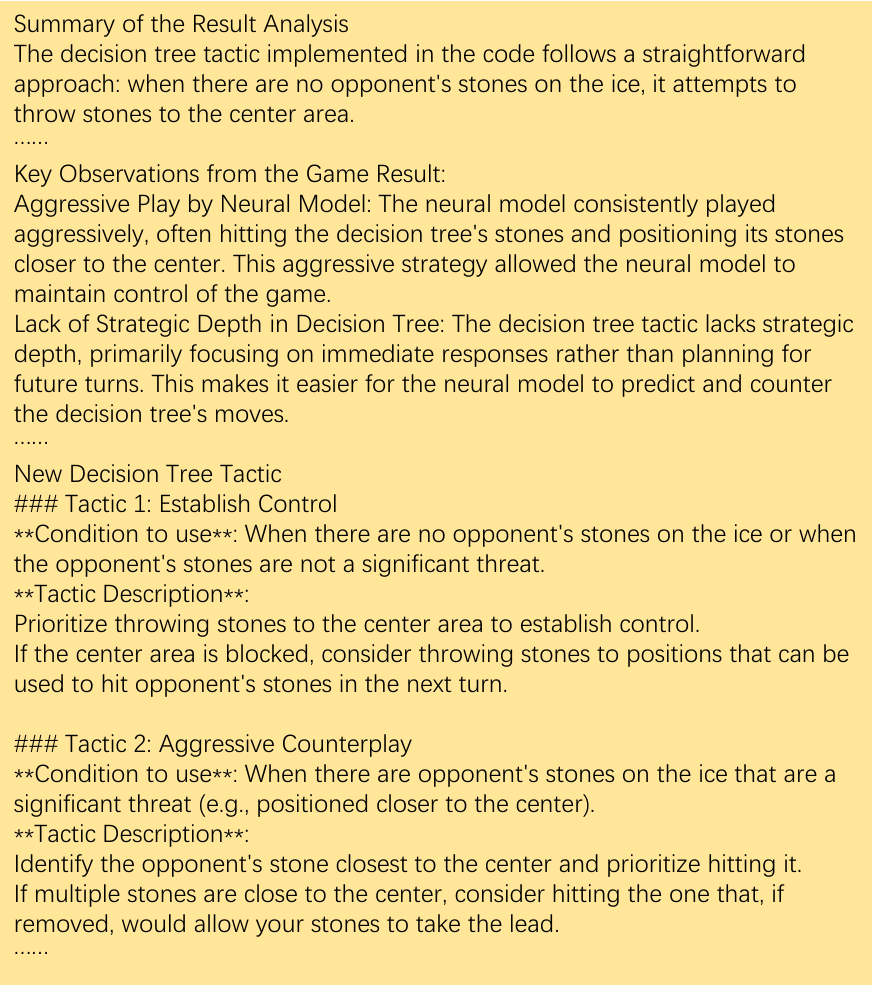}
	\caption{An LLM Critic output example.}
	\label{fig:criticanswer}
\end{figure}

The output shown in Figure~\ref{fig:criticanswer} reveals that the LLM Critic meticulously analyzes the shortcomings of the current decision tree and the strategies employed by the deep neural model to out-maneuver it. 
Drawing upon this analysis, the LLM Critic formulates an enhanced version of the decision tree.

\begin{figure}[t]
    \centering
    \begin{subfigure}[b]{0.45\textwidth}
        \includegraphics[width=\textwidth]{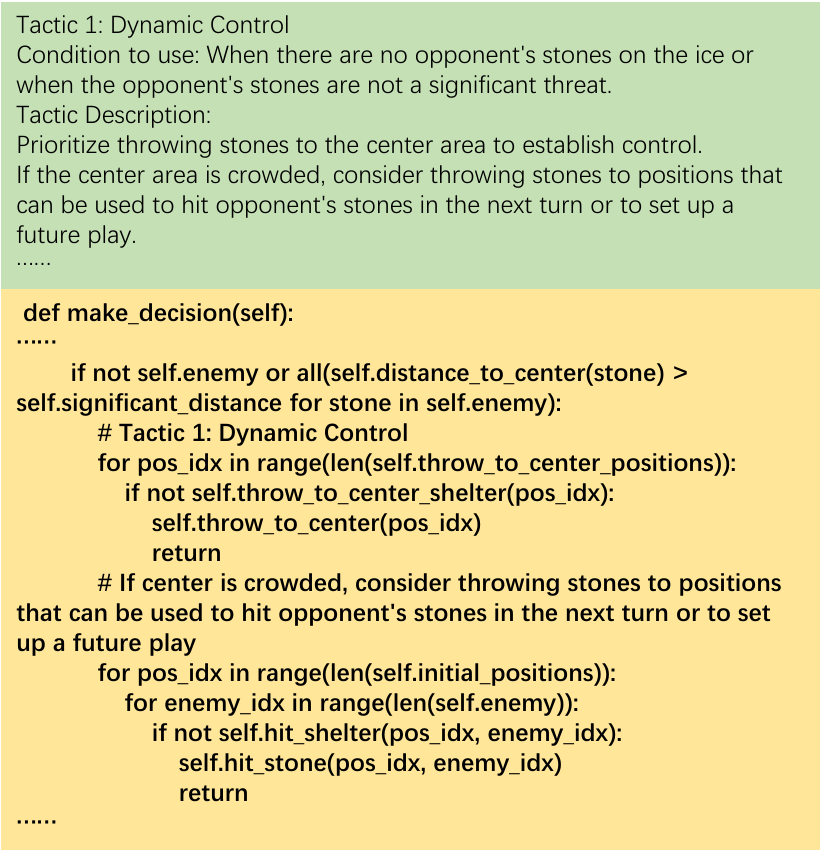}
        \caption{Decision Tree 3 Generation.}
        \label{fig:tree3}
    \end{subfigure}
    \hfill
    \begin{subfigure}[b]{0.45\textwidth}
        \includegraphics[width=\textwidth]{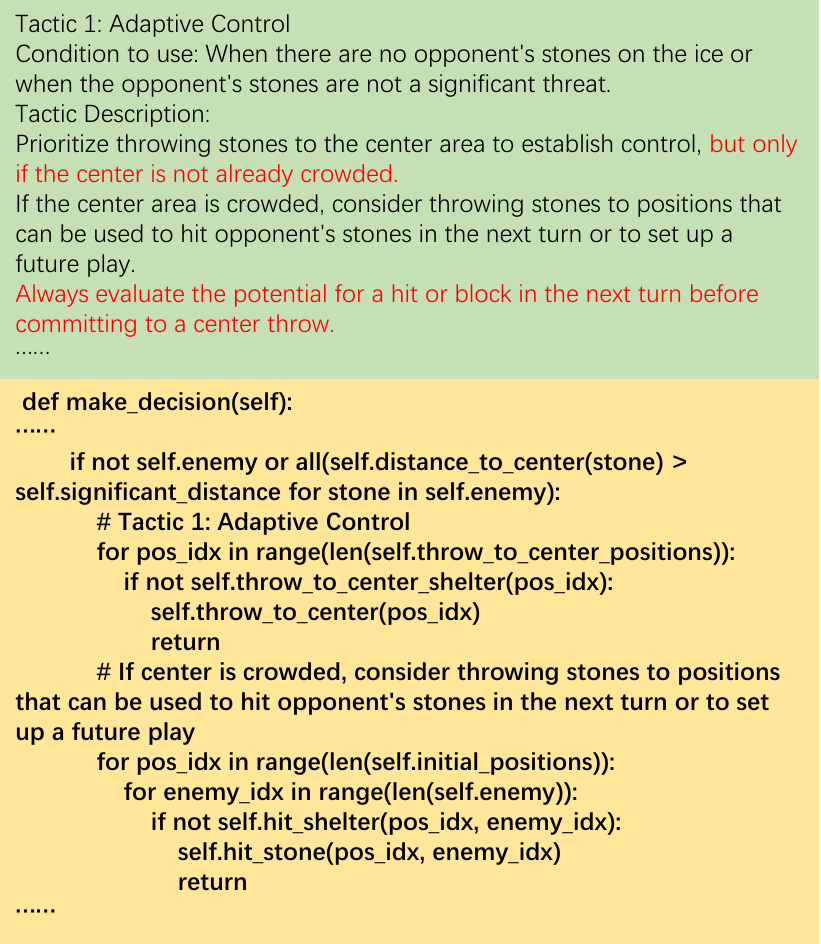}
        \caption{Decision Tree 4 Generation.}
        \label{fig:tree4}
    \end{subfigure}
    \caption{Decision Tree 3's and Decision Tree 4's Python codes are semantically identical.}
    \label{fig:contrast}
\end{figure}

Given that the curling game environment is not a widely recognized domain, the LLM Critic does not possess extensive prior knowledge about this specific environment. 
Consequently, after two cycles of refinement, when the decision tree ascends to the top rank on the Jidi platform, the LLM Critic's ability to provide clear and actionable improvement advice diminishes. 
As a result, the LLM Coder generates an identical decision tree code, with the only variation being the code comments, as shown in Figure~\ref{fig:contrast}.
In this way, we terminate the iterative refinement loop until the LLM Critic fails to further enhance the decision tree. 

\subsection{Discussion}

\subsubsection{Explainable Strategy}
The decision tree code generated by the LLM inherently constitutes an explainable strategy. 
In comparison to deep neural model policies derived from reinforcement learning, decision trees offer transparent and interpretable decision-making processes, enabling stakeholders to comprehend and trace the rationale behind each decision. 
This transparency is vital for fostering trust and accountability, particularly in critical applications such as autonomous driving.
Moreover, decision trees are inherently modular and can be effortlessly modified or extended to incorporate new rules or conditions, rendering them more adaptable to evolving environments or requirements. 
Conversely, deep neural network models often lack such interpretability and flexibility.
Furthermore, decision trees are less susceptible to overfitting compared to deep neural models, as they naturally capture hierarchical relationships. 
In contrast, policy models tend to perform poorly when competing against unseen opponents during training, highlighting the robustness and generalization ability of decision trees.

\subsubsection{Human Prior Knowledge}
The development of a decision tree inherently necessitates robust human prior knowledge, as it involves structuring rules and conditions that guide the decision-making process. 
This requirement presents a significant challenge, particularly in specialized environments such as the curling game, where expert knowledge may be scarce or inaccessible. 
However, large language models exhibit the remarkable capability to supply this essential knowledge to generate decision trees. 
By harnessing vast amounts of pre-existing text data, LLMs can distill pertinent information, identify patterns, and formulate logical structures that emulate human expertise. 
This capability not only mitigates the dependency on human experts but also accelerates the development process, enabling the creation of a sophisticated decision tree. 
Consequently, LLMs serve as invaluable tools in augmenting human knowledge and facilitating the efficient generation of decision trees, showing a potential to enhance the adaptability and robustness of AI systems in complex environments.

\section{Conclusion}	
In this paper, we have introduced a novel hybrid approach which leverages the strengths of both RL and LLMs to iteratively refine decision tree tactics, enhancing their performance and adaptability.
The proposed framework consists of three pivotal modules: the LLM Coder, the RL Debugger, and the LLM Critic. 
The LLM Coder generates executable Python code based on the specified decision tree tactic, ensuring consistency and accuracy in implementation. 
The RL Debugger trains a customized deep neural model to compete against the generated decision tree code, identifying its shortcomings.
The LLM Critic analyzes the loss trajectories and provides insights into the decision tree's vulnerabilities, generating an improved version of the tactic.
The iteratively refined decision tree curling AI achieves superior performance on the Jidi platform, ranking first among 34 competing teams.	

We posit that the advent of domain-specific Large Language Models presents a novel avenue for the generation of decision tree code, which can be subsequently refined through Reinforcement Learning to identify and help mitigate potential flaws. 
This integrated approach holds significant promise for the development of robust and high-performance strategies. 
Moving forward, we intend to extend the applicability of our methodology to other domains, including but not limited to chess, card games, and e-sports such as StarCraft, thereby broadening its impact and demonstrating its versatility across diverse strategic contexts.
	
	%
	
	\ifCLASSOPTIONcaptionsoff
	\newpage
	\fi

	
	
	%

{\bibliographystyle{IEEEtran}
\bibliography{mirror}}

\end{document}